# Validating a Deep Learning Algorithm to Identify Patients with Glaucoma using Systemic Electronic Health Records

John Xiang, BA[1]; Rohith Ravindranath, MS[1]; Sophia Y. Wang, MD, MS[1]

[1]Department of Ophthalmology, Byers Eye Institute, Stanford University, Stanford, California, USA

# Abstract

*We evaluated whether a glaucoma risk assessment (GRA) model trained on All of Us national data can identify patients at high probability of glaucoma using only systemic electronic health records (EHR) at an independent institution. In this cross-sectional study, 20,636 Stanford patients seen from November 2013 to January 2024 were included (15% with glaucoma). A pretrained GRA model was fine-tuned on the Stanford cohort and tested on a held-out set using demographics, systemic diagnoses, medications, laboratory results, and physical examination measurements as inputs. The best model achieved AUROC 0.883 and PPV 0.657. Calibration was consistent with clinical risk: the highest prediction decile showed the greatest glaucoma diagnosis rate (65.7%) and treatment rate (57.0%). Performance improved with more trainable layers up to 15 and with additional data. An EHR-only GRA model may enable scalable and accessible pre-screening without specialized imaging.*

# Introduction

Many medical conditions remain underdetected because screening and early diagnosis may require specialized examinations and equipment that are not routinely available in general medical settings.[1–3] A practical informatics need is to use routinely collected electronic health record (EHR) data to stratify risk and prioritize who should receive downstream specialty evaluation, while ensuring that models generalize across institutions and patient populations.[4,5] This is particularly important because clinical prediction models often degrade when deployed in new settings due to differences in coding practices, patient mix, and measurement processes (dataset shift), making external validation central to responsible translation.[4,5]

Glaucoma is a representative, high-impact example: it is a leading cause of irreversible blindness, is frequently undiagnosed until late stages, and shows persistent demographic disparities in burden and detection.[1,6,6–9] Typical screening and confirmatory assessment rely on optic nerve evaluation and imaging, which require specialized ophthalmic equipment that is not routinely available in primary care settings.[1] As a result, identifying who should be prioritized to receive ophthalmic evaluation for glaucoma remains challenging, and there is no consensus on an appropriate screening population. The United States Preventive Services Task Force (USPSTF) has highlighted the need for further research on risk assessment tools to identify individuals at higher risk who may benefit most from targeted screening.[1]

We have developed predictive models applying deep learning techniques to EHR data to identify patients with high risk of developing glaucoma from their systemic health information, without the need for specialized ophthalmic examinations, aimed at facilitating early efficient and scalable detection of glaucoma.[10,11] These models were trained on EHR data from participants in the All of Us Research program, an initiative spearheaded by the National Institutes of Health to collect clinical, environmental, lifestyle, and genetic data from more than 850,000 individuals in the US.[12] The best performing model was a deep learning approach that combined autoencoders with a 1-dimensional convolutional neural network, performing well in predicting glaucoma diagnosis with an AUROC of 0.863 and a positive predictive value (PPV) of 0.587. These results indicated that almost 60% of patients the model predicted as high risk were coded with glaucoma diagnoses. This corresponds to substantial enrichment of recorded glaucoma diagnoses in the model-positive group compared with population prevalence (typically <5%), suggesting potential utility as an EHR-based glaucoma risk-assessment or pre-screening step to prioritize downstream ophthalmic evaluation.[13] Use of such an EHR-based glaucoma risk assessment model to prioritize patients for photography-based glaucoma screening may improve screening efficiency by enriching the tested population relative to baseline prevalence.

An important next step toward translation is external validation in an independent health system and assessment of whether EHR-predicted glaucoma risk aligns with clinically meaningful glaucoma-associated measures such as intraocular pressure and cup-to-disc ratio, which are not readily available in All of Us. However, because EHR feature distributions and coding practices vary across health systems, practical deployment will likely require site-specific adaptation. In this study we evaluate both transportability of the All of Us-pretrained model and the extent of fine-tuning needed to recover performance in an independent eye center cohort. We also examine whether model-predicted risk aligns with clinical glaucoma indicators available in the eye center EHR, including intraocular pressure and cup-to-disc ratio.

# Methods

*Data Source and Cohort*
Our cohort comprised 20,636 adult patients who received care at the Stanford Byers Eye Clinic between November 2013 and January 2024, who had at least one intraocular pressure (IOP) measurement and a non-ophthalmology encounter at Stanford prior to their first recorded IOP date. EHR data for each patient was extracted from Stanford's clinical data warehouses: systemic data was extracted in the Observational Medical Outcomes Partnership (OMOP) Common Data Model (CDM) format, which utilizes the same data schema as the All of Us dataset, and ophthalmology-specific eye examination data was extracted separately as representation in OMOP for these data elements is incomplete.[14] Glaucoma was defined as requiring at least two encounters with an associated glaucoma diagnosis of any type, excluding glaucoma suspects, as determined by OMOP concept codes. This study was approved by the Stanford Institutional Review Board (IRB) and adhered to the Declaration of Helsinki.

*Data Preprocessing and Feature Engineering*
Input features to the glaucoma risk assessment (GRA) model are identified and preprocessed the same way as for the original AoU model. This consisted of demographic data (age, sex, race/ethnicity), encounter diagnoses and medications stored as Boolean variables, and laboratory measurements and physical exam findings considered continuous values that are standardized. Additionally, the earliest date of glaucoma diagnosis was identified for each glaucoma patient so only systemic features prior to that diagnosis were used as inputs to the model. In addition to these model input features, we also extracted additional eye-specific features that were used to evaluate the model performance, including the maximum recorded IOP across both eyes, the maximum recorded cup-to-disc ratio (CDR) across both eyes, and information on prescribed glaucoma medications, selective laser trabeculoplasty (SLT), glaucoma surgeries and other lasers in both eyes. We imputed missing values using scikit-learn's iterative chained-equations approach, a widely used implementation of multivariate imputation by chained equations (MICE) for multivariable missingness.[15–17] The dataset was divided into training, validation, and test sets in a 70/10/20 ratio (Ns = 14,445/2063/4128) at the patient level.

*Model Architecture, Inference, and Experiments*
We evaluated transportability by applying the pre-trained All of Us model to the Stanford cohort and evaluated adaptability under dataset shift using transfer learning via fine-tuning. We utilized the same model architecture as the original study, which included two autoencoders designed to transform one-hot encoded medication and diagnosis data into dense feature representations. These representations were then concatenated with the rest of the input features and fed into a 1-dimensional convolutional neural network (1D-CNN) to create dense feature matrices for diagnoses and medications. This experimental setup is illustrated in **Figure 1**. The GRA model was implemented in TensorFlow 2.11.0 and scikit-learn v1.2.1 was used for processing and analysis.[16,18]

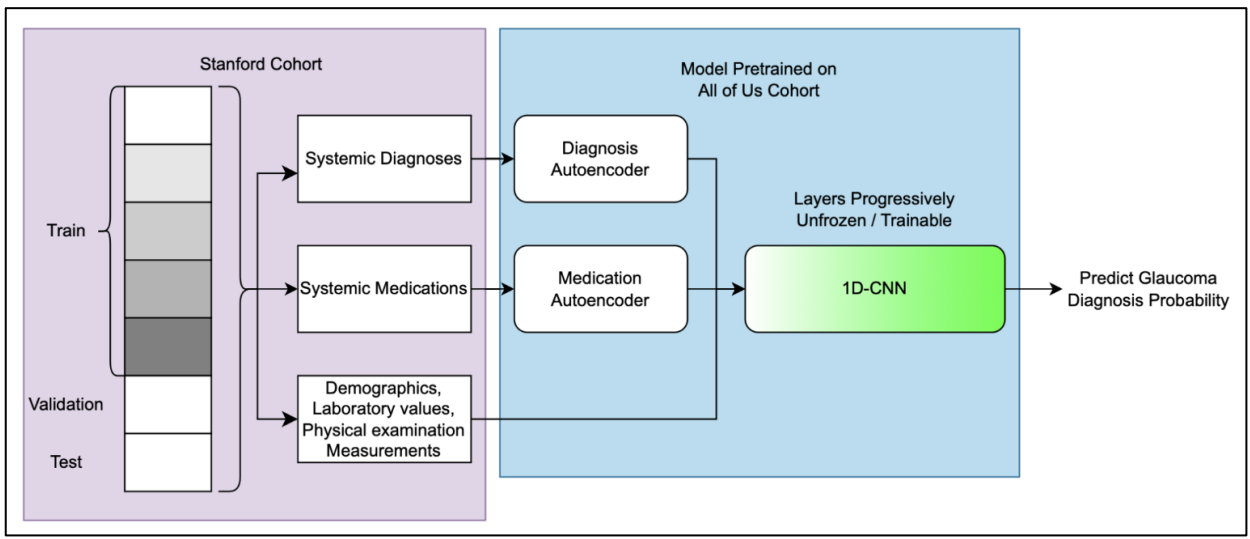

**Figure 1.** Experimental setup showing Stanford cohort data going through the pretrained diagnosis and medication autoencoders with the trainable 1D-CNN.

To investigate the performance of the AoU-pre-trained model on the Stanford cohort, we conducted experiments to determine the optimal number of layers of the 1D-CNN to freeze while still achieving high performance. Freezing all layers would prevent the model from adapting to the new dataset at all during the fine-tuning process, while unfreezing all layers (making all layers trainable) could lead to the model unlearning existing knowledge, learning too much new information and potentially overfitting to the new dataset. While freezing the pre-trained autoencoders for diagnosis and medication representation, we varied the number of frozen vs trainable layers in the final 1D-CNN and compared performance metrics to identify the best configuration for transfer learning. When varying this number, the difference in number of layers between experiments was not consistently 1 because not every layer had trainable parameters. Additionally, the unfreezing began from the final layers of the model and moved towards the input layer (e.g. started with the last layer unfrozen, then the last two layers, etc.) because the back/output layers typically capture task-specific features while the initial layers encode more general representations.

We also conducted experiments to evaluate how varying the amount of Stanford training data affected model performance, aiming to determine whether accurate predictions could still be achieved with smaller cohorts, such as those likely encountered when deploying the model in new settings with limited data on hand. We systematically varied the amount of training data in increments of 20% while keeping the test set constant and compared the performance metrics to better understand the relationship between the quantity of training data and the model's ability to generalize effectively. As a baseline comparator, we trained an XGBoost classifier using only demographic variables (age, sex, race/ethnicity), applying the same train/validation/test splits and preprocessing (e.g., encoding/scaling) as the GRA models.[19] We report its performance alongside the AoU-pretrained and fine-tuned variants to provide a simple demographics-only reference, as glaucoma risk is known to vary according to demographic characteristics.

*Evaluation*

As each model outputs a decimal classification score between 0 and 1, we tuned a binary classification boundary for every model based on maximizing F1 score on the validation set prior to test evaluation. On the held-out test set, we calculated AUROC, area under the precision-recall curve (AUPRC), accuracy, recall, precision, and F1 score (the harmonic mean of recall and precision). Additionally, we evaluated the model using both standard calibration curves and similarly styled risk stratification plots on glaucoma-related features (e.g., maximum IOP, CDR). In our calibration curves, predicted probabilities were bucketed into deciles, and the average predicted risk in each group was plotted on the x-axis against the actual observed glaucoma prevalence on the y-axis. Calibration curves are used in predictive modeling to assess how well the predicted probabilities are aligned with actual observed outcomes, providing insight into the reliability of the model's risk estimates.[20] Ophthalmic variables used for calibration analysis such as glaucoma medications, lasers, surgeries, IOP, and CDR were excluded from model inputs and used only for evaluation.

# Results

**Table 1** describes the demographic characteristics of our study cohort, which included a total of 20,636 patients who received eye care at Stanford, among whom 3,165 (15.34%) had glaucoma and 17,471 (84.66%) did not have glaucoma. The mean age of the cohort was 64.64 years old, with a standard deviation of 18.42 years. The majority of the cohort was female, comprising 58.14% (N=11,997) of the patients. Non-Hispanic White people constituted around half of the population at 50.13% (N=10,345), with a sizable Asian population making up over a quarter at 26.24% (N=5,414).

**Table 1.** Characteristics of the study cohort.

|  | Overall Mean | Overall SD | Glaucoma Mean | Glaucoma SD | Non-Glaucoma Mean | Non-Glaucoma SD |
|---|---|---|---|---|---|---|
| Age | 64.64 | 18.42 | 75.34 | 15.30 | 62.70 | 18.28 |
|  | Total Population | Total Population % | Glaucoma Patients | Glaucoma Patients % | Non-Glaucoma Patients | Non-Glaucoma Patients % |
| N | 20636 | 100.00% | 3165 | 100.00% | 17471 | 100.00% |
| Male | 8637 | 41.85% | 1377 | 43.51% | 7260 | 41.55% |
| Female | 11997 | 58.14% | 1788 | 56.49% | 10209 | 58.43% |
| Non-Hispanic White | 10345 | 50.13% | 1356 | 42.84% | 8989 | 51.45% |
| Non-Hispanic Black | 726 | 3.52% | 171 | 5.40% | 555 | 3.18% |
| Non-Hispanic | 5414 | 26.24% | 1059 | 33.46% | 4355 | 24.93% |

| Asian | | | | | | |
| Hispanic | 2405 | 11.65% | 335 | 10.58% | 2070 | 11.85% |
| Other Race/Ethnicity | 1746 | 8.46% | 244 | 7.71% | 1502 | 8.60% |

*Model Performance*

We fine-tuned and evaluated the performance of the All of Us model on the Stanford cohort, while systematically varying the size of the dataset used for fine-tuning and the number of trainable layers. The relationship between model performance and the number of trainable layers and the size of the fine-tuning dataset is shown in **Figure 2**. In general, increasing the number of trainable layers and/or the size of the fine-tuning dataset improved the model performance (**Figure 3**). **Figure 4** shows the receiver operating characteristic curves of the best model, which achieved AUROC of 0.883 and AUPRC of 0.692. This model had 15 trainable layers and utilized the full 100% of the Stanford cohort training dataset. The performance of the models stratified by race/ethnicity is also shown, with the highest AUROC observed in Non-Hispanic White (AUROC 0.892) and the highest AUPRC observed in Non-Hispanic Black (AUPRC 0.771); across groups, AUROC ranged from 0.834 to 0.892 and AUPRC ranged from 0.611 to 0.771.

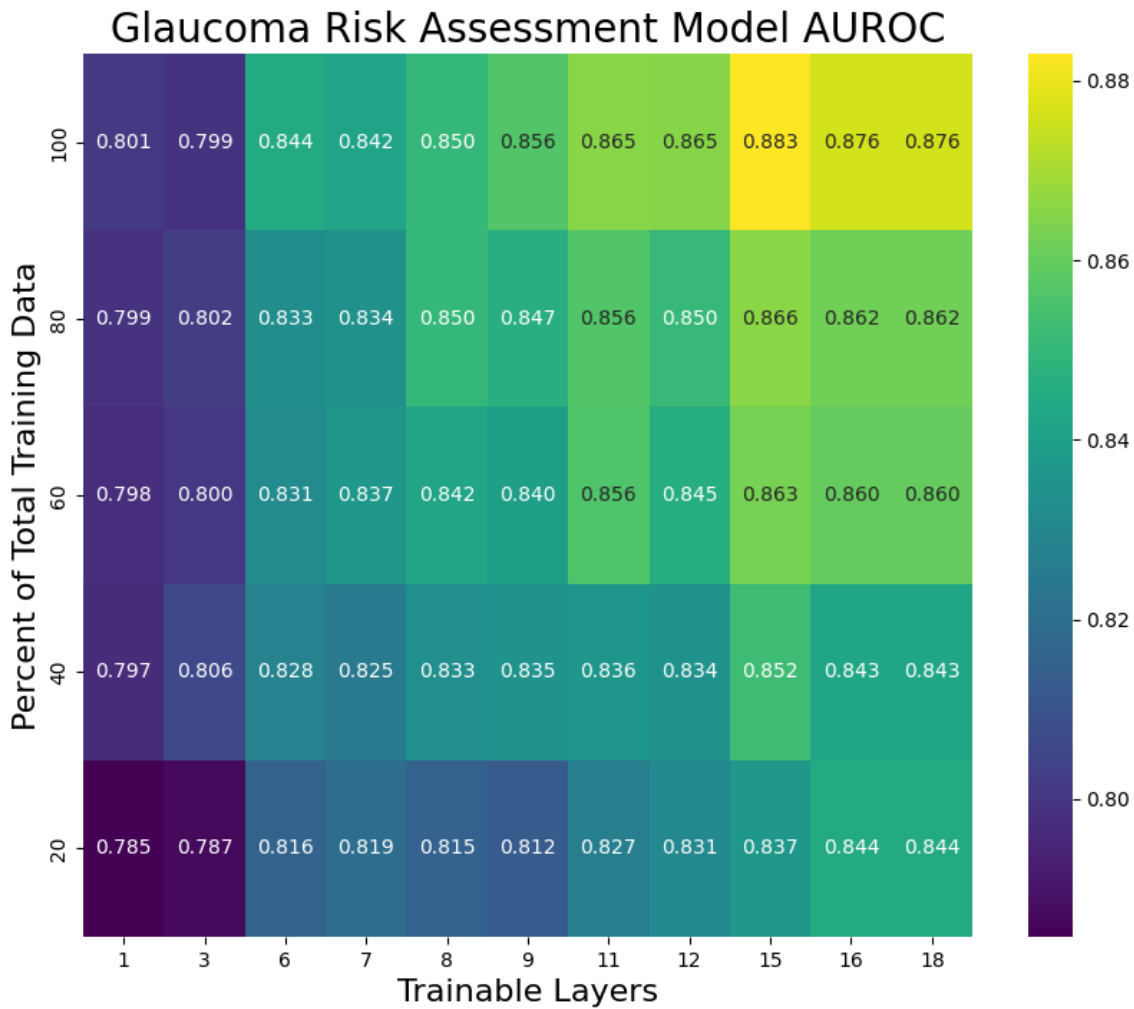

**Figure 2.** Heatmap showing the relationship between model performance, number of trainable layers, and amount of training data. The x-axis represents the number of trainable layers in the model, increasing from left to right. The y-axis represents the percentage of the training dataset used, increasing from bottom to top. Warmer colors indicate better performance measured by area under the receiver operating characteristic curve (AUROC). Model performance improves toward the upper-right corner, suggesting that both increased amounts of training data and greater model flexibility contribute to better performance on the external test set.

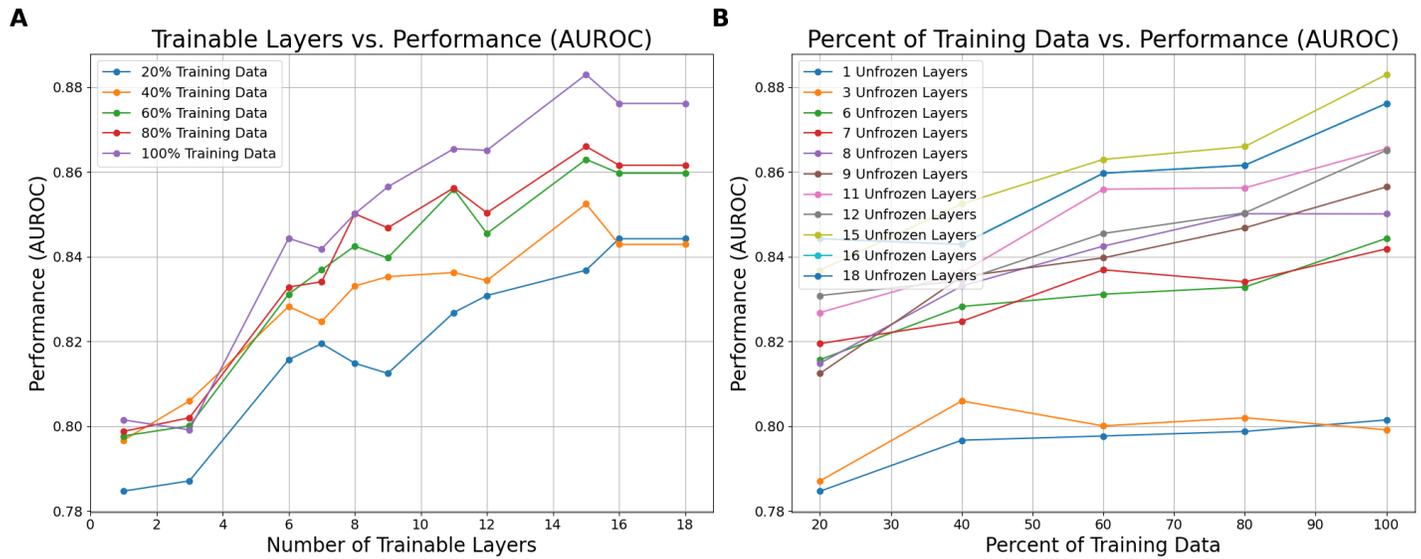

**Figure 3.** Model performance vs size of the training set and number of trainable layers. (A) plots area under the receiver operating characteristic curve (AUROC) as a function of the number of trainable layers with the size of the training data held constant. (B) plots AUROC as a function of the fraction of training data used with the number of trainable layers held constant.

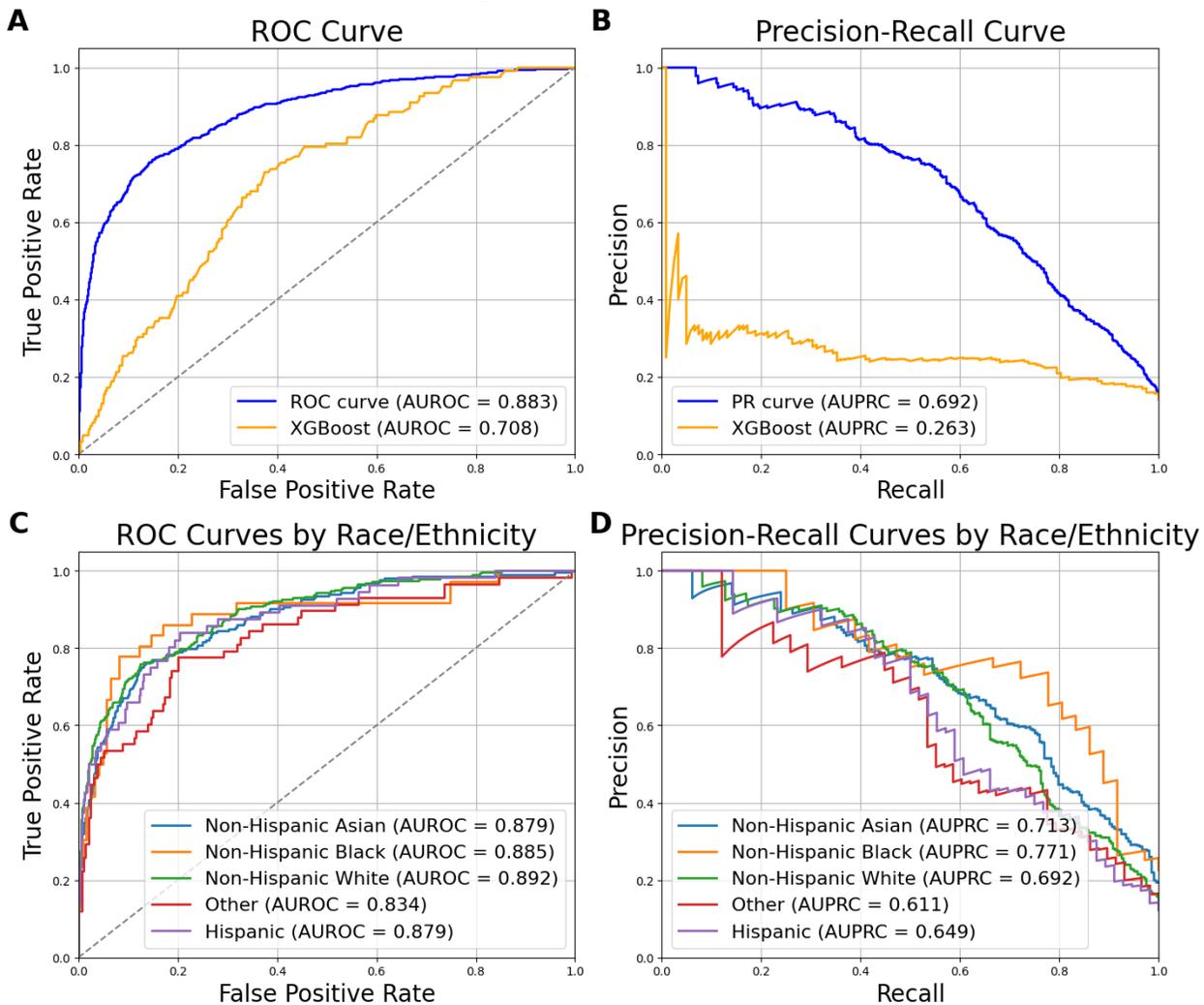

**Figure 4.** Area under the receiver operating and precision recall curves for best-performing model. (A) shows the receiver operating characteristic (ROC) curve for the best model and the baseline XGBoost model, while (B) shows their precision-recall curves. (C) shows the best model's ROC curve stratified by race/ethnicity and (D) shows its precision-recall curve stratified by race/ethnicity.

This model also demonstrated superior performance across other classification metrics (e.g. AUPRC, sensitivity), showing similar performance to those models with 16 and 18 trainable layers, with AUROC values 0.876 and above. These models also used a majority of the Stanford training set. The model with all layers frozen did not perform well at an AUROC of 0.467, well below the rest of the results which ranged from 0.801 to 0.883. Full performance metrics are shown in **Table 2**. As a reference, the demographics-only XGBoost baseline (age, sex, race/ethnicity) achieved AUROC 0.708 and AUPRC 0.263 on the test set, and it performed worse than the fine-tuned GRA models.

**Table 2.** Performance metrics for the best model per number of trainable layers.

| Trainable Layers | Percentage | AUROC | AUPRC | Accuracy | Sensitivity | Specificity | PPV (Precision) | NPV | F1 Score | Best Threshold |
|---|---|---|---|---|---|---|---|---|---|---|
| 0 | 40 | 0.467 | 0.141 | 0.156 | 1.000 | 0.000 | 0.156 | 0.000 | 0.271 | 0.00 |
| 1 | 100 | 0.801 | 0.523 | 0.809 | 0.594 | 0.849 | 0.422 | 0.919 | 0.494 | 0.60 |
| 3 | 40 | 0.806 | 0.504 | 0.772 | 0.692 | 0.786 | 0.375 | 0.932 | 0.487 | 0.60 |
| 6 | 100 | 0.844 | 0.611 | 0.861 | 0.548 | 0.919 | 0.556 | 0.916 | 0.552 | 0.70 |
| 7 | 100 | 0.842 | 0.619 | 0.847 | 0.618 | 0.889 | 0.508 | 0.926 | 0.558 | 0.70 |
| 8 | 80 | 0.850 | 0.615 | 0.868 | 0.539 | 0.929 | 0.584 | 0.916 | 0.560 | 0.85 |
| 9 | 100 | 0.856 | 0.642 | 0.883 | 0.492 | 0.955 | 0.669 | 0.910 | 0.567 | 0.90 |
| 11 | 100 | 0.865 | 0.668 | 0.871 | 0.638 | 0.914 | 0.580 | 0.932 | 0.608 | 0.80 |
| 12 | 100 | 0.865 | 0.662 | 0.884 | 0.577 | 0.941 | 0.643 | 0.923 | 0.608 | 0.85 |
| **15** | **100** | **0.883** | **0.692** | **0.889** | **0.610** | **0.941** | **0.657** | **0.929** | **0.632** | **0.90** |
| 16 | 100 | 0.876 | 0.687 | 0.883 | 0.596 | 0.936 | 0.632 | 0.926 | 0.614 | 0.80 |
| 18 | 100 | 0.876 | 0.687 | 0.883 | 0.596 | 0.936 | 0.632 | 0.926 | 0.614 | 0.80 |

Calibration curves for the best performing model are shown in **Figure 5**. These plot the actual rate of glaucoma diagnosis, the maximum recorded IOP, the maximum CDR, and the rate of glaucoma treatment with medication, selective laser trabeculoplasty, or surgery, against deciles of model-predicted glaucoma risk. As the model-predicted glaucoma risk increases, all of these factors rise in a similar manner. The calibration curves rise steeply at approximately the top 1 to 3 deciles of glaucoma risk. For example, the actual rate of glaucoma diagnosis in the top decile of model-predicted glaucoma risk is 65.7%; the rate of glaucoma treatment in this group is 57.0%, the average maximum recorded IOP is 22.326 mmHg, and the average maximum CDR recorded is 0.651. In contrast, the lowest values were observed across the lower deciles including 1.8% for diagnosis, 6.4% for treatment, 17.320 mmHg for maximum IOP, and 0.316 for maximum CDR, indicating a clear risk gradient. Normal IOP is defined as 10–21 mmHg per the American Academy of Ophthalmology, and population studies report a mean CDR near 0.40 in non-glaucomatous adults, with values around 0.60 or higher often considered suspicious for glaucoma.[21–23]

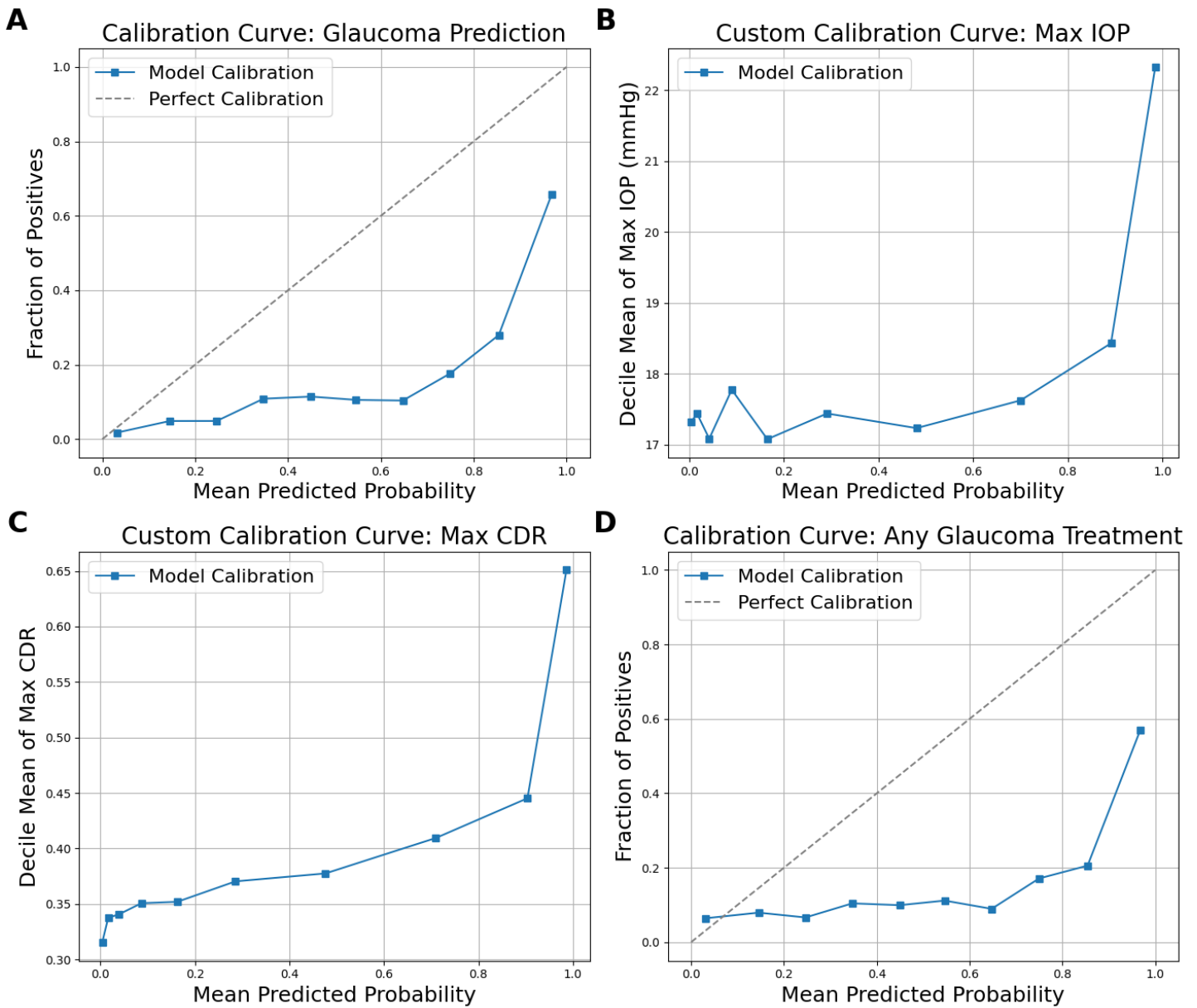

**Figure 5.** Calibration curves for model validation against glaucoma diagnosis codes and clinical parameters. (A) shows a standard calibration curve where the x-axis corresponds to the model's predicted probability of glaucoma grouped by decile and the y-axis showing the actual observed glaucoma frequency for that bucket. (B) plots the models' predicted glaucoma probability against maximum recorded intraocular pressure (IOP). (C) does the same for maximum recorded cup-to-disk ratio (CDR). (D) plots against records of any glaucoma treatment, including glaucoma medication, laser, or surgery. For the binary outcomes, a dashed diagonal line indicating perfect calibration, meaning predicted probabilities precisely match the observed frequencies, is also plotted for comparison.

# Discussion

In this study, we demonstrated that an EHR-based deep learning model for glaucoma risk assessment trained on a national cohort can be transported to a local independent health system using transfer learning within a shared OMOP data model, recovering strong discrimination and showing coherent calibration to clinical indicators. Glaucoma is a leading cause of blindness and half of patients with glaucoma do not know they have it, making it an ideal use case for evaluating an EHR-only pre-screening system.[6,24,25] With fine-tuning on the Stanford cohort, the model achieved an AUROC of 0.883 and a PPV of 0.657, which is comparable to the original model's performance (AUROC 0.863, PPV 0.587).[11] Additionally, the model maintained strong performance in different race/ethnicity groups, as the AUROC performance ranged from 0.834 to 0.892. These results suggest that this modeling approach can identify glaucoma risk in new patient populations, even in the absence of ophthalmic data in the input features, but fine-tuning the model is necessary. Validation on an external dataset is particularly important given the common challenge of model generalizability in clinical AI, and highlights the necessity of testing models across diverse populations to ensure their robustness and real-world utility.[26,27]

We explored the effects of varying the amount of training data and the number of trainable layers in the 1D convolutional neural network (CNN) part of the model during fine-tuning, while keeping the medication and diagnosis autoencoders frozen. Our findings indicated that increasing the proportion of training data generally improved model performance, with diminishing returns observed beyond approximately 80% of the training dataset (N=11,556). Additionally, allowing more layers to be trainable during fine-tuning enabled the model to better adapt to the new dataset, suggesting that deeper retraining may be beneficial when transferring models across clinical populations. However, as our best model did not have the maximum number of trainable layers (15 out of a possible 18), there was a balance to be found in leaving some of the layers frozen and retaining more of what the model had learned from the larger All of Us dataset on which it was originally trained in addition to the frozen autoencoders.

A major strength of this study was in detailed calibration curve analysis, not only against glaucoma diagnoses in the EHR but also against a variety of clinical parameters related to glaucoma that were not available in the original All of Us cohort. Calibration analysis reveals whether higher model-predicted scores are actually indicative of higher risk of glaucoma in the test cohort. In this case, those in the highest decile of model-predicted glaucoma risk had over 35 times higher rate of actual glaucoma diagnosis (65.7% vs 1.8% rate of glaucoma). Similarly, the model's output scores also aligned with other clinically meaningful features, such as maximum IOP and CDR: those with greater model-predicted risk of glaucoma also had much higher maximum recorded IOP and CDR. These clinical features were available in our Stanford dataset but not in the original All of Us training data, making this analysis uniquely valuable for assessing the model's extended utility. There was a threshold effect observed across the calibration curves, where glaucoma rates and clinical parameters increase sharply in the highest deciles of predicted risk. Thus, in future applications using these models to identify high-risk patients for glaucoma screening, there may be a natural threshold for referring those patients.

From a broad informatics perspective, scalable decision support should leverage routinely collected EHR data and interoperable standards such as the OMOP CDM, while being evaluated for performance under dataset shift across settings.[28,29] In contrast, much of today's clinical AI remains imaging-first, with strong performance but recurring workflow and integration hurdles. For example, Hussain et al. developed a multimodal deep learning framework that integrates optical coherence tomography (OCT) images, visual field data, and clinical variables to predict glaucoma progression, while Wu et al. developed a model for intracranial hemorrhage that attained AUROC 0.96 and PPV ~86–89% on internal and external sets and matched neuroradiologists on a 100-case subset while being considerably faster per scan.[30,31] While these studies highlight the potential of imaging-based AI, a key limitation in applicability for screening and diagnosis is imaging patients at scale, which can require costly equipment, trained staff, standardized acquisition protocols, and patient scheduling, creating access, cost, and heterogeneity barriers. Another study examining multi-site qualitative deployment of an imaging-based chest-CT assistant found additional challenges with site-to-site variability, integration with existing systems, and privacy concerns.[32] By contrast, our EHR-only method avoids specialized imaging integrations and adapts locally via light fine-tuning. Deploying specialized imaging at scale can be challenging without first identifying high-risk individuals who warrant advanced imaging, which is where the utility of approaches like ours which do not require such specialized data shines.

Another widely used disease risk-stratification approach is the polygenic risk score (PRS), which can be constructed with statistical or machine learning methods and provides a genomics-based signal that complements clinical informatics models across diseases.[33–37] In cardiovascular medicine, multi-ancestry PRS have improved risk discrimination and reclassification for coronary artery disease and can help target primary-prevention decisions such as statin initiation in younger adults with borderline or intermediate clinical risk.[38,39] Within ophthalmology, Singh et al. incorporated a PRS into a model for predicting primary open-angle glaucoma (POAG) onset in participants from the Ocular Hypertension Treatment Study, finding that the addition of PRS as a covariate significantly improved baseline model performance.[34] Their analysis using similar calibration-style curves also showed that higher PRS values were associated with increased POAG risk, with a clear positive trend in their calibration-style plot: the highest PRS decile corresponded to a 21.81% conversion rate to POAG, with the lowest at 9.52%. In context, our model's highest risk decile achieved a 65.7% positive rate for glaucoma diagnosis, with the lowest at 1.8%. These findings reinforce the potential of EHR features as a tool for early identification and targeted monitoring, perhaps even in combination with PRS in the future. While it may be more difficult to perform glaucoma pre-screening with PRS than with using routinely-collected EHR data due to the logistical, financial, and infrastructural barriers of genotyping a general population, nevertheless, PRS and AI models may be able to work in a complementary manner. PRS can provide an input factor representing a stable, lifelong genetic risk baseline, while the AI models could reflect evolving clinical data.

We acknowledge several limitations in this study. First, our validation was conducted at a single academic center, which may limit the generalizability of our findings to other healthcare settings. Future work will involve external validation across additional institutions, including those within the Observational Health Data Sciences and Informatics (OHDSI) network or other sites with EHR data standardized to the OMOP CDM, which is increasingly being adopted as a standard for observational health data. Studies across the OHDSI network can further assess model generalizability across diverse populations and environments. Model interpretability poses another limitation because the architecture uses autoencoders for high-dimensional EHR features, which constrains direct feature attribution; future work could pair this approach with simplified, attribution-friendly models for explanation. In addition, the models were developed and validated in eye clinic populations in whom patients had a chance to have their glaucoma diagnosed and thus, glaucoma prevalence was relatively high; future studies will also evaluate performance in a true screening population where patients may not have previous eye care and the prevalence of glaucoma is likely to be lower. Separately, another limitation was the use of billing codes to determine glaucoma diagnosis, as billing codes can be an imperfect proxy due to variability in coding practices and potential misclassification. To partially address this, we supplemented our evaluation with calibration-style analyses using related clinical features, which provided additional evidence of the model's predictive validity. Finally, while our calibration analyses suggest the model captures meaningful clinical signals, further work is needed to understand how these predictions translate across different subpopulations, including those with varying comorbidities, socioeconomic backgrounds, or healthcare access patterns. Prospective studies should also evaluate stability over time and assess whether locally set thresholds are needed as prevalence and practice evolve.

# Conclusion

This study illustrates an informatics pathway for transporting and locally adapting an EHR-based risk stratification model across health systems. Starting from a large national cohort, we fine-tuned the model on an independent site and recovered strong discrimination; performance improved with more local training data and a greater number of trainable layers, with diminishing returns near the upper range. Calibration-style risk stratification aligned higher predicted risk with higher observed outcome prevalence and with domain indicators available in the local EHR (diagnosis codes, treatment rates, maximum IOP and CDR) supporting use for prioritization and decision support. Together, these results motivate the use of EHR for disease prescreening as a scalable, noninvasive component of clinical AI pipelines and underscore the importance of transfer learning under dataset shift, common-data-model–aligned feature spaces, and site-specific adaptation for deployment. Future work will include prospective evaluation, evaluation in true screening populations, threshold selection under operational constraints, and multi-site validation.